\documentclass[lettersize,journal]{IEEEtran}
\usepackage{graphicx}
\usepackage{amsmath,amsfonts}
\usepackage{algorithm}
\usepackage{array}
\usepackage{subfigure}
\usepackage{textcomp}
\usepackage{stfloats}
\usepackage{url}
\usepackage{verbatim}

\usepackage{cite}
\usepackage{xcolor}
\usepackage{bm}
\usepackage{listings}
\lstdefinestyle{promptstyle}{
    basicstyle=\ttfamily\footnotesize,
    breaklines=true,
    breakatwhitespace=false,
    columns=fullflexible,
    keepspaces=true,
    frame=single,
    numbers=none,
    numberstyle=\tiny,
    captionpos=b,
    xleftmargin=1.5em,
    framexleftmargin=1.2em
}

\usepackage{graphicx}
\usepackage{multirow}
\usepackage{amssymb}
\usepackage{amsmath}
\usepackage[colorlinks=true, citecolor=green, linkcolor=red]{hyperref}  
\usepackage{algorithm}
\usepackage{orcidlink}
\usepackage[noend]{algpseudocode}
\begin{document}

\title{CoLMIN: LLM-based Multi-Decision Path Negotiation for Cooperative Autonomous Driving}

\author{ 
    Zhe Huang\orcidlink {0009-0005-7656-1298},  Zhaoxin Fan\textsuperscript{*}\orcidlink{0000-0002-6324-1712}, Shuo Wang\orcidlink{0000-0002-6720-1646},  Wenjun Wu\orcidlink{0000-0003-2998-8828}, Xuan Zhao\orcidlink{0000-0003-0119-7768}, Min Liu\orcidlink{0000-0001-6406-4896}
\thanks{
Zhe Huang is with the School of future Transportation, Chang'an University, Xi'an, China (e-mail: huangzhe21@chd.edu.cn) 

Zhaoxin Fan  is with the Hangzhou International Innovation Institute, Beihang University, Beijing, China (Email: zhaoxinf@buaa.edu.cn)

Wenjun Wu is with the Hangzhou International Innovation Institute, Beihang University, Beijing, China (Email: wwj09315@buaa.edu.cn)

Shuo Wang is with the School of information, Renmin University of China, Beijing, China, 100872 (e-mail:shuowang18@ruc.edu.cn) 

Xuan Zhao is with the School of Automobile, Chang 'an University, Xi'an, China  (e-mail:zhaoxuan@chd.edu.cn)

Min Liu is with the School of Artificial Intelligence and Robotics,
Hunan University, and the National Engineering Research Center for Robot Visual Perception and Control Technology, Changsha, China (e-mail:liu\_min@hnu.edu.cn)

}

}


\markboth{Accepted for Publication in IEEE Transactions on Multimedia}%
{Shell \MakeLowercase{\textit{et al.}}: A Sample Article Using IEEEtran.cls for IEEE Journals}



\maketitle

\begin{abstract}

Multi-vehicle cooperative autonomous driving enhances the safety and reliability of autonomous driving systems through information sharing among connected vehicles, demonstrating significant potential for improving traffic safety.
LLM-based approaches leverage strong reasoning capabilities of LLMs to enable effective inter-vehicle negotiation and improve cooperative driving performance.
However, driving decisions in complex traffic scenarios are inherently multi-solution in nature.
As a result, existing negotiation-based methods often converge prematurely to suboptimal solutions, hindering consensus formation and limiting the practical deployment of cooperative autonomous driving systems. 
To address this challenge,  we propose CoLMIN, the LLM-based multi-decision path negotiation framework for cooperative autonomous driving, achieving stable decision consensus through multi-decision  path negotiation and reflective reasoning. 
\color{black}
\color{black}
To achieve stable and high-quality consensus in cooperative autonomous driving, CoLMIN consists of three key components: (i) an LLM-based Multi-Intent Negotiation module (LMin), which adopts a Negotiator–Evaluator paradigm and generates multiple candidate driving intentions for joint evaluation,
(ii) an Evaluation-based Shallow Reflection Module (ESRM), which analyzes negotiation outcomes and provides feedback to guide subsequent negotiations, thereby accelerating consensus formation; and (iii) an LLM-based Deep Reflection Module (LDRM) that performs long-term reflection over negotiation histories to mitigate cognitive fixation and prevent the system from converging to suboptimal solutions.
Experimental results in CARLA simulation environment demonstrate that CoLMIN significantly outperforms existing methods in challenging interactive driving scenarios. \color{black}The code is released at \url{https://github.com/HuangZhe885/CoLMIN}.\color{black}

\end{abstract}

\begin{IEEEkeywords}
Collaborative Perception, Autonomous Driving, Large Language Models, Multi-agent consensus, Multi-agent  negotiation
\end{IEEEkeywords}

\section{Introduction}

\begin{figure}[t]
\centering
\includegraphics[width=1\columnwidth]{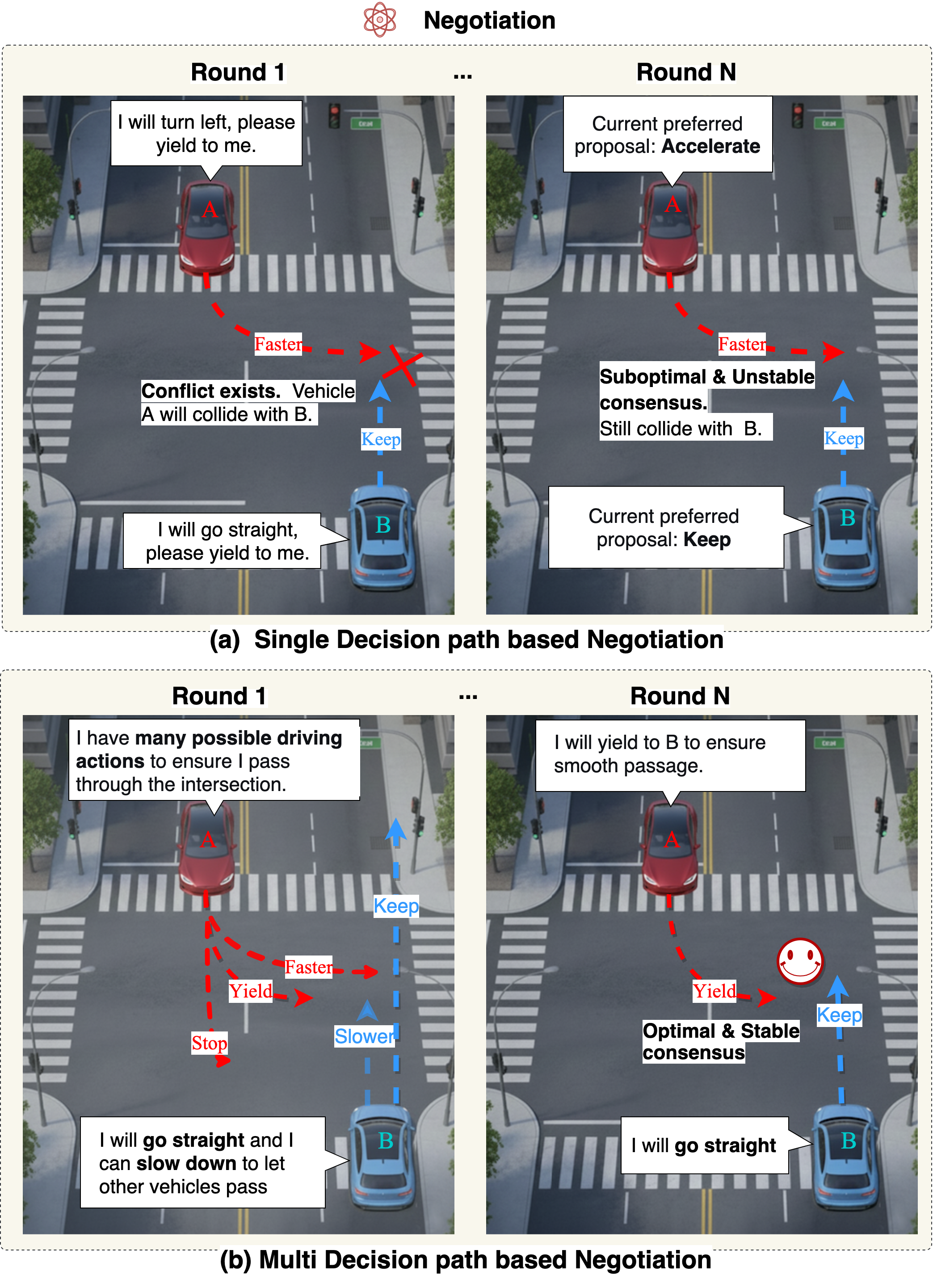} 
\caption{\color{black}Consensus quality under different negotiation paradigms. CoLMIN adopts a multi-decision path based negotiation paradigm to generate and evaluate negotiation outcomes, achieving optimal and stable consensus.\color{black}}
\label{Fig-Negotiate}
\end{figure}

\IEEEPARstart{R}ecent advances in cooperative autonomous driving have demonstrated that multi-agent collaboration can significantly enhance the safety and efficiency of traffic systems, thereby attracting considerable attention from the research community \cite{CoDrivingLLM,CoDriving,lmdrive}. 
Unlike single-vehicle autonomous driving systems \cite{huang,tmm1,tmm2,tmm3}, which rely solely on their own observations to make driving decisions, cooperative autonomous driving leverages real-time information sharing and intention communication \cite{huang2,s2,cui2026expert}. Such cooperative approaches enable collaborative environmental perception among multiple vehicles.
As a result, the inherent limitations of single-vehicle driving, including safety risks caused by incomplete perception and uncertainty in surrounding vehicles’ behaviors, can be effectively mitigated.

Traditional cooperative autonomous driving systems can be generally divided into three categories: optimization-based methods, rule-based methods, and learning-based methods. Optimization-based methods \cite{o2,o3,o1,o4} formulate multi-vehicle planning as a constrained optimization problem. Although they are straightforward to design, they often require task-specific objectives and constraints, suffer from high computational complexity, and struggle to handle unseen scenarios. In contrast, rule-based methods  \cite{r1,r2,r3} are characterized by simplicity and computational efficiency, but their reliance on predefined rules often results in limited robustness. Learning-based methods  \cite{UniAD,LAV,TCP} have been successfully applied to various cooperative autonomous driving tasks and have demonstrated remarkable performance. However, most of existing learning-based models often experience significant performance degradation when transferred to unseen environments \cite{l4,tmm5}, highlighting the need for more flexible and interactive cooperative autonomous driving frameworks.

Recently, Large Language Models (LLMs) have demonstrated remarkable reasoning capabilities and a vast knowledge base, opening new opportunities for cooperative autonomous driving systems \cite{llm1,tmm4}. 
By leveraging the reasoning and commonsense knowledge of LLMs, cooperative autonomous driving systems can negotiate at a semantic level, thereby enhancing operational efficiency and safety in complex traffic environments.
Specifically, compared with traditional cooperative autonomous driving approaches, LLM-based methods are able to understand diverse and dynamic traffic scenarios, facilitate sophisticated decision-making \cite{colmdriver,CoDrivingLLM}, and reduce communication overhead through language-mediated interaction \cite{v1,s4}, thereby providing greater flexibility. 
LangCoop \cite{v1} is a representative work of this class of methods, which proposes an collaborative autonomous driving paradigm that leverages natural language as a compact yet expressive medium for inter-agent communication, striking a balance between communication efficiency and semantic integrity. 
Building upon this paradigm, CoDrivingLLM \cite{CoDrivingLLM} introduces an interactive and learnable cooperative decision-making framework that enables vehicles to negotiate through language, thereby achieving adaptive cooperative autonomous driving across diverse traffic scenarios.
To enable real-time driving control, CoLMDriver further \cite{colmdriver} presents the first full-pipeline cooperative autonomous driving system, which incorporates negotiated high-level intentions into low-level control modules, enabling language-driven real-time cooperative autonomous driving control.

\color{black}Although recent studies have made significant progress in LLM-based cooperative autonomous driving\color{black}, we observe that most current methods still follow a single-decision path based negotiation paradigm, in which each agent generates only one deterministic driving plan in each negotiation round, and the system updates consensus solely around this single set of plans. 
However, this paradigm exhibits two critical issues in real-world traffic scenarios.
First, it overlooks the inherent multi-solution nature of driving decision-making. Under the same traffic conditions, there often exist multiple feasible and safe driving strategies. 
Restricting negotiation to a single plan may cause the system to prematurely converge to a suboptimal solution. 
Second, when negotiation is conducted solely around a single plan proposed by each agent, the process becomes confined to repeated comparisons of deterministic decisions. In this situation, agents are more likely to develop overconfidence in their own decisions and tend to adhere to their original plans, 
They are often unwilling to make substantive adjustments during negotiation, which can lead to negotiation deadlock and ultimately increase the risk of consensus failure.
\color{black}
\color{black}

Fig. \ref{Fig-Negotiate} illustrates the differences between single-decision and multi-decision path based negotiation paradigms in a typical intersection scenario. 
As shown in Fig. \ref{Fig-Negotiate}(a), under the single-decision path based negotiation paradigm, vehicle A plans to turn left while vehicle B intends to go straight, and both vehicles expect the other to yield.
Since each vehicle generates only one deterministic action in each negotiation round, the negotiation process is confined to discussing these fixed decisions. Consequently, both vehicles become confident in the correctness of their own actions and are unwilling to revise their initial plans, causing the negotiation to stagnate. 
In contrast, Fig. \ref{Fig-Negotiate}(b) presents a multi-decision path based negotiation scenario, where vehicles A and B are able to simultaneously consider multiple feasible and safe driving plans. By jointly discussing these alternatives, vehicles can flexibly adjust their strategies during negotiation, thereby alleviating overconfidence in initial decisions and facilitating the formation of a globally optimal and stable consensus.

\color{black}
Motivated by the above observation, we propose \textbf{CoLMIN} (Cooperative LLM-based Multi-decision path Negotiation), a cooperative autonomous driving framework that leverages LLMs to enable multi-decision path generation, negotiation and reflective reasoning.
Different from existing LLM-based cooperative driving methods, CoLMIN explicitly introduces multi-decision-path negotiation to better capture the multi-solution nature of driving decisions and facilitate stable decision consensus.
\color{black}
The proposed framework allows cooperative autonomous driving systems to move beyond the single-decision path based negotiation paradigm by generating multiple candidate driving intents and selecting the optimal driving solution through multi-round negotiation. 
Specifically, to achieve stable and high-quality consensus in multi-vehicle negotiation, CoLMIN is designed to consist of three key components: \color{black}
(1) a LLM-based Multi-Intention Negotiation Module (LMin).
This module is designed to establish a stable negotiator–evaluator mechanism, enabling each vehicle to generate multiple feasible candidate intentions within a single negotiation round and thereby expanding the negotiation space.
(2) an Evaluation-based Shallow Reflection Module (ESRM),
which targets local convergence difficulties during the negotiation process. By analyzing negotiation outcomes, ESRM produces improvement suggestions and feeds them back into the subsequent negotiation rounds, accelerating the gradual convergence of consensus. 
(3) a LLM-based Deep Reflection Module (LDRM),
which aims to mitigate overconfidence and reasoning fixation that gradually emerge during negotiation. By analyzing historical records and feedback across multiple negotiation rounds, LDRM explicitly identifies and corrects long-term cognitive biases, enhancing strategy diversity.

To evaluate the performance of CoLMIN, we conducted extensive experiments across ten challenging traffic scenarios.
The experimental results demonstrate that CoLMIN significantly outperforms existing cooperative autonomous driving methods, demonstrating much higher  success rates across all scenarios and establishing new state-of-the-art performance. Notably, we find that CoLMIN is significantly superior in complex traffic scenarios.
In the challenging lane-changing (LC) scenario, CoLMIN improves safety by 8\% and increases the overall success rate by 13\%, demonstrating its effectiveness and robustness under complex traffic environments.

Our contributions can be summarized as:

\begin{itemize}
\item  We propose CoLMIN, an LLM-based multi-decision path negotiation framework for cooperative autonomous driving, enabling stable and robust decision consensus in complex traffic scenarios. 
\item  We design three key components that jointly address multi-decision path generation, consensus acceleration, and overconfidence mitigation, namely the LMin module, the ESRM module, and the LDRM module, enabling robust and stable consensus formation in multi-vehicle negotiation. 
\item We conduct comprehensive experiments across  complex traffic scenarios, demonstrating that the proposed CoLMIN method achieves state-of-the-art performance in metrics such as safety, efficiency, and success rate.
\end{itemize}

\section{Related works}
\subsection{End-to-end Autonomous Driving}
End-to-end autonomous driving has emerged as a major research topic, aiming to map raw environmental observations directly to control signals. Existing approaches can be categorized into reinforcement learning based methods and imitation learning based methods.
Reinforcement learning \cite{tmm6,LR2,LR3} constructs an interactive training framework where the autonomous agent progressively acquires navigation and control capabilities through continuous environmental interaction mechanisms.
Imitation learning primarily aim to reproduce expert agent behaviors by fitting recorded driving data, representing a key research direction in end-to-end autonomous driving. Several representative methods, such as  TransFuser \cite{transfuser}, ReasonNet \cite{reasonnet} and InterFuser \cite{InterFuser}, adopt Transformer-based architectures to capture fine-grained scene representations. Other methods provide additional learning signals to enhance the effectiveness of policy learning. For example, TCP \cite{TCP} synchronizes trajectory and control predictions to improve behavior cloning by intermediate distillation, while LAV \cite{LAV} learns driving policies from both experiences of the ego and surrounding vehicle.
UniAD \cite{UniAD} integrates full-stack driving tasks—including tracking, mapping, motion prediction, occupancy forecasting, and planning—into a unified network, connecting all modules through a query-based design.

Although these methods improve the accuracy of end-to-end autonomous driving, they exhibit limited generalization and reasoning capabilities when encountering unseen scenarios. To address these issues, we propose a framework that leverages LLMs to generate potential driving plans and employs a multi-decision path based negotiation paradigm to achieve optimal and stable consensus decisions.


\subsection{LLMs-based Driving}

With the integration of LLMs, autonomous driving systems have gradually evolved toward unified architectures that combine perception, reasoning, and decision-making, improving system interpretability and enabling human-like interaction capabilities \cite{llm2,llm3,llm5}. 
Several studies jointly process multi-modal sensor inputs and natural language instructions to generate driving actions. 
DriveLM \cite{drivelm} integrates vision–language models (VLMs) pretrained on web-scale data into end-to-end driving systems to enhance generalization ability, while VLM-AD \cite{vlmad} leverages VLMs as teacher models to provide additional supervision during training. 
In parallel, a number of datasets have been constructed to study question-answering (QA) tasks in autonomous driving. 
HAD \cite{HAD} focuses on human–vehicle interaction guidance, Lingo-QA \cite{lingoqa} designs counterfactual QA tasks to evaluate reasoning and decision-making in complex traffic scenarios.
Despite these advances, most existing LLM-based driving studies primarily focus on enhancing single-vehicle autonomy, while research on multi-vehicle cooperation remains relatively limited. 
LangCoop \cite{v1} employs natural language as an inter-vehicle communication medium to reduce communication overhead during information exchange. 
CoDrivingLLM \cite{CoDrivingLLM} introduces a roadside-unit-based negotiation scheme for resolving inter-vehicle conflicts but does not produce executable vehicle control commands. 
To bridge this gap, CoLMDriver \cite{colmdriver} proposes a  framework with an explicit waypoint generator, enabling the translation of negotiation outcomes into real-time driving control signals.

\color{black}
Although existing methods improve reasoning and decision-making in autonomous driving, the single-decision path based negotiation paradigm often restricts the exploration of alternative feasible strategies, causing the system to converge prematurely to suboptimal plans and repeatedly revise decisions, thereby hindering consensus formation.
As a result, incorporating reflection mechanisms into the negotiation process becomes essential for improving convergence behavior and enhancing consensus quality.

\color{black}
\subsection{LLM-based negotiation}

Negotiation, as a fundamental mechanism for resolving conflicts and achieving mutually beneficial outcomes, has been widely studied in disciplines such as game theory \cite{game}, economics \cite{g2}, and psychology \cite{g3}. With the rapid development of multi-agent systems and LLMs, recent studies have increasingly explored LLM-based negotiation frameworks \cite{r3,r4}. Representative works include GENTEEL-NEGOTIATOR \cite{r1}, which combines LLMs with a Mixture-of-Experts (MoE) reinforcement learning framework to improve negotiation quality and adaptability, and Interact, Instruct to Improve \cite{r2}, which proposes a parallel Actor-Reasoner architecture to reconcile reasoning capability with real-time constraints.

However, existing methods are primarily designed for general negotiation dialogues or human-agent interaction tasks, with a focus on language generation, strategy articulation, and interaction quality. They do not adequately address the challenges inherent in cooperative autonomous driving, such as dynamic conflicts and consensus formation among multiple vehicles.  As a result, we proposes an LLM-based negotiation framework for autonomous driving, which facilitates intention exchange, conflict resolution, and consensus formation among multiple vehicles, thereby enhancing the safety and reliability of cooperative decision-making in dynamic traffic environments.
\color{black}

\subsection{Multi-agent consensus}
In recent years, a growing body of research on multi-agent systems explores how multiple agents can collaborate through debate and negotiation to solve complex reasoning tasks \cite{con1,Divthink,FactReason,camel}.
Compared with single-agent systems, such collaborative paradigms demonstrate unique advantages in reasoning and decision-making under complex conditions.
For example, CAMEL \cite{camel} introduces a role-playing communication framework that guides conversational agents to jointly accomplish tasks through prompt engineering, RoCo \cite{RoCo} equips each robot with a language model to enable strategic discussion and collective reasoning, FactReason-MAD \cite{FactReason} designs a consensus-seeking task in which agents negotiate to reach a shared conclusion, and DivThink-MAD \cite{Divthink} develops a multi-agent debate framework that encourages agents to express opposing views under the supervision of a judge agent to derive the final solution.
Similarly, ChatEval \cite{chateval} constructs a team of judge agents to autonomously discuss and evaluate model responses in open-ended tasks.

Although  these multi-agent negotiation frameworks have shown promising results in general reasoning tasks, their application to cooperative autonomous driving remains challenging due to the stringent safety of dynamic traffic environments. 
In cooperative autonomous driving, limited mutual understanding of other vehicles’ intentions often leads to decision conflicts and instability during negotiation, which may increase collision risks. In contrast, 
\color{black}
CoLMIN introduces a multi-decision-path-based negotiation paradigm for cooperative autonomous driving, explicitly accounting for the multi-solution nature of driving decisions and facilitating more stable and effective consensus formation.
\color{black}

\section{Problem Formulation}
\label{formulation}
\color{black}
We consider a cooperative autonomous driving system consisting of $N$ connected autonomous vehicles operating in a shared traffic environment. Each vehicle $i$ aims to reach its destination $D_i$ while coordinating its behavior with other vehicles. Each vehicle perceives its local environment through onboard sensors. The raw sensor data are processed by the perception module and transformed into high-level semantic representations. We use $X_i$ to denote the semantic perception information of vehicle $i$. Vehicles interact with each other through communication, sharing their respective perception information and initial driving intentions $\{X_i, I_i\}_{i=1}^{N}$.

In this work, we model LLM-based cooperative autonomous driving as a constrained iterative consensus optimization problem, where the LLM serves as a negotiation module to analyze, evaluate, and coordinate the driving decisions of multiple vehicles in a bounded iterative process. The objective of CoLMIN is to find a joint driving strategy that satisfies safety constraints while improving efficiency and consensus quality. Formally, the goal can be expressed as:
\begin{align}
\arg \max_{\theta} \sum_{i=1}^{N} d\bigl( \Phi_{\theta}(X_i, D_i, I_i) \bigr),
\label{eqn:overall}
\end{align}
where $d(\cdot)$ denotes the driving performance metric, and $\Phi_{\theta}$ represents the LLM-based negotiation framework parameterized by $\theta$.
Specifically, at negotiation round \(r\), the LLM generates a finite set of candidate negotiation plans:
\begin{align}
\mathcal{P}^{(r)} = \{p^{(r)}_1, p^{(r)}_2, \ldots, p^{(r)}_{K_r}\}, 
\quad r =0, 1,2,\dots,R_{\max},
\end{align}
where \(p^{(r)}_m\) denotes the \(m\)-th candidate negotiation plan generated in round \(r\), and \(K_r\) is the number of candidate plans. Each candidate plan represents a complete multi-vehicle negotiation suggestion, including the driving actions, interaction requests, and responses of the participating vehicles.

For each candidate plan \(p^{(r)}_m\), the evaluator computes the overall negotiation score \(S_{score}(p^{(r)}_m)\), which is used to evaluate the quality of the candidate plan. The best candidate in round \(r\) is selected as:
\begin{align}
\hat{p}^{(r)} = 
\arg\max_{p^{(r)}_m \in \mathcal{P}^{(r)}} 
S_{score}(p^{(r)}_m).
\end{align}

To support bounded iterative optimization, we define the best-so-far candidate up to round \(r\) as:
\begin{align}
p^{*,r}
=
\arg\max_{p \in \bigcup_{t=1}^{r}\mathcal{P}^{(t)}}
S_{score}(p),
\end{align}
and the corresponding best-so-far score as:
\begin{align}
S_r^* = S_{score}(p^{*,r}).
\end{align}

The negotiation terminates when
\begin{align}
S_r^* \ge \tau \quad \text{or} \quad r = R_{\max},
\end{align}
where \(\tau\) is a predefined negotiation threshold and \(R_{\max}\) is the maximum number of negotiation rounds.
\color{black}
 
\section{Methodology}
This section introduces CoLMIN, an LLM-based multi-decision path negotiation framework for cooperative autonomous driving.
We first present an overview of the overall system architecture in Section~\ref{Overall}, followed by detailed descriptions of the LLM-based Multi-Intent Negotiation Module (LMin), the evaluation-based Shallow Reflection Module (ESRM), and the LLM-based Deep Reflection Module (LDRM) in Sections~\ref{LMin}, \ref{ESRM}, and \ref{DRM}, respectively.

\subsection{Overall Architecture}
\label{Overall}

\begin{figure*}[t]
\centering
\includegraphics[width=2\columnwidth]{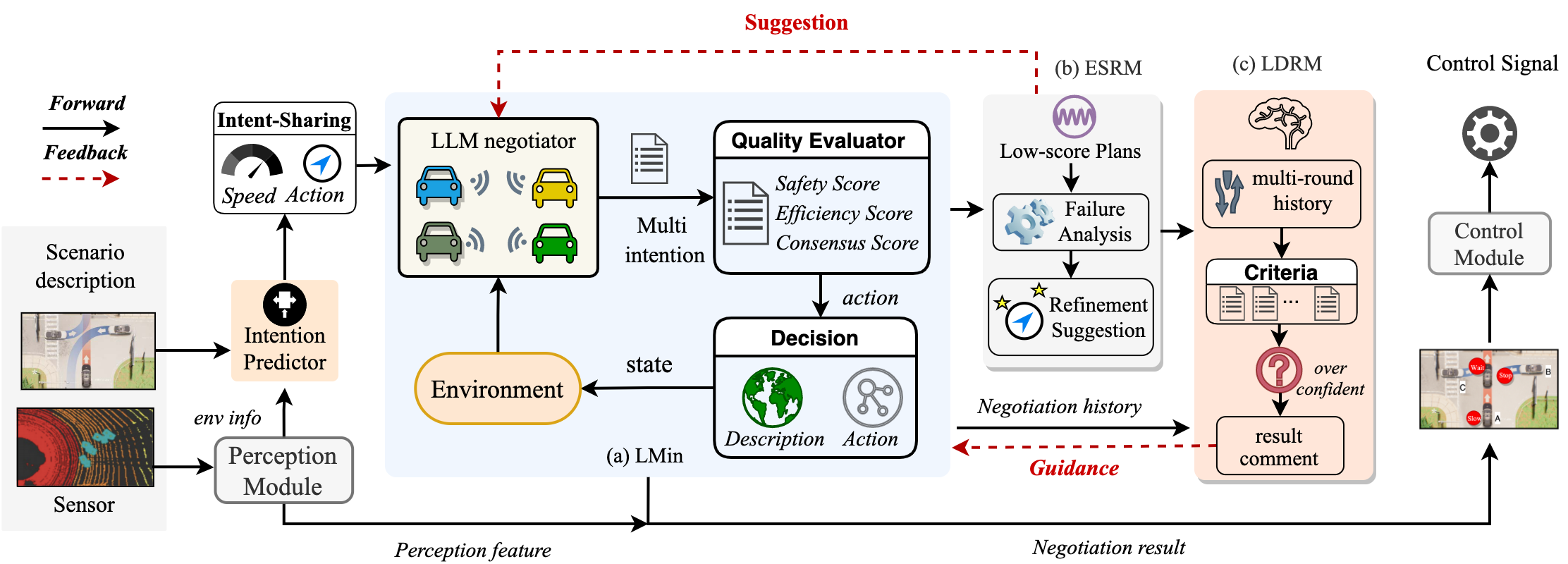} 
\caption{ \color{black}
Overall architecture of CoLMIN. The framework consists of the perception module and three negotiation-related modules. The perception module converts raw sensor data into environmental information for intention detection. When intention conflicts are detected, LMin performs multi-vehicle negotiation. ESRM provides short-term refinement to accelerate consensus, while LDRM uses long-term reflection to mitigate overconfident or repetitive decisions. \color{black}
}
\label{Fig-overall}
\end{figure*}

\color{black}
As shown in Fig.~\ref{Fig-overall}, in CoLMIN, upon receiving sensor data, the perception module generates intermediate representations, including object-level 3D information, BEV perception features, and BEV occupancy representations. Among them, the BEV perception features and BEV occupancy maps are used as auxiliary inputs for planning tasks, while the object-level perception results and interaction-related information are converted into concise textual descriptions.
The perception outputs and textual descriptions are fed into an intent predictor following the architecture in \cite{colmdriver}, which infers each vehicle’s initial driving intention and predicts whether potential conflicts exist. The inferred intention encodes the vehicle’s current motion state, including its speed and intended driving action.
\color{black}

When conflicts are detected, the initial driving intentions of all vehicles are transmitted to the LMin module. 
\color{black}
Specifically, LMin first generates and evaluates multiple feasible candidate driving plans according to safety, efficiency, and consensus related criteria. 
If the candidate plans generated in the current round fail to satisfy the predefined quality requirements, the ESRM module is triggered to identify low-quality plans and provide short-term refinement suggestions. These suggestions are fed back to LMin to guide candidate plan revision in the next negotiation round.
When the negotiation proceeds for multiple rounds, the LDRM module further analyzes the negotiation history and adjusts the strategy to avoid repetitive or overconfident decisions. Through the iterative interaction among LMin, ESRM, and LDRM, CoLMIN progressively refines the candidate plans and returns a stable consensus driving plan within bounded negotiation rounds.
 \color{black}
The final negotiated plan is then passed to the low-level control module, which converts the high-level decision into executable vehicle control signals, such as steering, throttle, and braking.

\subsection{LLM-based Multi-Intent Negotiation Module}
\label{LMin}

\color{black}
The LLM-based Multi-Intent Negotiation module aims to establish a stable Negotiator–Evaluator mechanism to generate multiple feasible candidate driving intentions, thereby expanding the negotiation space and avoiding premature convergence to suboptimal decisions.

\textbf{Overall Process.}
The negotiation proceeds over multiple rounds and follows the Negotiator–Evaluator paradigm. As illustrated in Fig.~\ref{Fig-Negotiate}, the negotiator engages in language-based negotiation with conflicting vehicles and generates multiple candidate driving intentions. These candidates represent different safe and traffic-compliant driving actions.
\color{black}
The input to the negotiator includes the ego vehicle’s speed and initial intention, the states and intentions of other vehicles, and feedback suggestions from the evaluator from the previous round, when available.
\color{black}
Based on the dialogue in the current round, the evaluator aggregates the candidate intentions of all vehicles and evaluates them from multiple perspectives. If none of the candidate plans satisfies the predefined criteria, the negotiation proceeds to the next round, where the negotiator proposes refined alternatives. This process continues until a feasible and consistent plan is identified.
Once consensus is reached, the final negotiated driving intention is passed to the downstream control module for execution. 
In summary, the LMin module consists of two tightly coupled components: a negotiator, which generates multiple candidate intentions through interaction, and an evaluator, which assesses these candidates and guides the negotiation toward consensus.

\textbf{Negotiator.} 
The negotiator module generates multiple executable solutions through language-based negotiation, thereby expanding the negotiation space. 
Specifically, in each negotiation round, participating vehicles “speak” in a predefined order. Each vehicle integrates the broadcast information from other vehicles and any available external feedback, and sequentially proposes multiple feasible solutions.
In the initial negotiation round, no external feedback is provided. In subsequent rounds, feedback from the ESRM and LDRM modules is incorporated to refine the proposed solutions and guide the negotiation toward better outcomes. The outputs of the negotiator include the vehicle’s own driving actions, requests to other vehicles, or responses to their behaviors.

\color{black}
Formally, at negotiation round \(r\), the LLM generates a finite set of candidate negotiation plans $\mathcal{P}^{(r)}$.
\color{black}
The generation process can be written as:
\begin{align}
\mathcal{P}^{(r)}
=
\mathrm{LLM}_{\theta}
\left(
f\left(
\{X_i,I_i\}_{i=1}^{N},
\mathcal{H}^{(r-1)},
F_{\mathrm{sugg}}^{(r-1)}
\right)
\right),
\end{align}
where \(f(\cdot)\) is the prompt generation function, \(\mathcal{H}^{(r-1)}\) denotes the negotiation history before round \(r\), and \(F_{\mathrm{sugg}}^{(r-1)}\) denotes the feedback suggestions from the previous round. In the first round, \(F_{\mathrm{sugg}}^{(0)}=\varnothing\).

As the inference time of LLMs increases with output length, the negotiator uses concise prompts to support efficient information exchange. The number of candidate plans, the speaking order, and the prompt format are fixed across experiments rather than tuned separately for individual scenarios.
\color{black}

\textbf{Evaluator.}
The evaluator module is responsible for assessing the candidate negotiation plans generated by the negotiator and providing feedback for the next negotiation round. \color{black} After receiving the candidate set \(\mathcal{P}^{(r)}\), the evaluator parses each candidate plan \(p_m^{(r)}\) into a joint driving intention representation for all participating vehicles.

For each candidate plan, the corresponding joint intention is distributed to the vehicles. Each vehicle then uses the low-level planning module to generate future waypoints according to the received intention. Therefore, the evaluator does not rely only on textual negotiation results, but also uses the generated future waypoints to assess the feasibility and quality of each candidate plan.
For each candidate plan \(p_m^{(r)}\), the evaluator computes the overall negotiation score as:
\begin{align}
S_{score}(p_m^{(r)})
=
w_S S(p_m^{(r)})
+
w_E E(p_m^{(r)})
+
w_C C(p_m^{(r)}),
\end{align}
where \(S(\cdot)\), \(E(\cdot)\), and \(C(\cdot)\) denote the safety, efficiency, and consensus scores, respectively. The safety score evaluates whether the planned trajectories are collision-free. The efficiency score measures whether the vehicles can make sufficient progress along their intended routes. The consensus score measures whether the vehicles are willing to execute the proposed joint strategy. \color{black}
The safety and efficiency scores are computed based on the future waypoints generated by the planner, \color{black} while the consensus score is generated by the LLM under a fixed evaluation protocol with predefined scoring criteria and standardized input representations. 
Specifically, the consensus scoring adopts structured prompts, standardized inputs, and a fixed output format. Based on the predefined scoring criteria, the LLM outputs a brief analysis and the corresponding consensus score. \color{black}
The candidate selection, best-so-far update, and termination rule follow the formulation defined in the \ref{formulation} section.

Unlike \cite{colmdriver}, which primarily uses the evaluation score for classification, our method leverages the total score to control the negotiation loop. If the score of a candidate plan exceeds a predefined threshold $\tau$, the negotiation process terminates and the plan is adopted as the consensus strategy. Otherwise, negotiation continues, with the highest-scoring plan selected in each round. 
This evaluation strategy ensures that even if consensus is not reached within a predefined number of negotiation rounds, the system can still provide the best feasible driving plan under the current traffic conditions.

\subsection{Evaluation-based Shallow Reflection Module}
\label{ESRM}
The Negotiator-Evaluator mechanism provides a stable framework for consensus formation by iteratively generating and evaluating candidate plans. However, in practical negotiation processes, convergence is not always guaranteed.
Specifically, in a given negotiation round, the negotiator may generate multiple candidate plans, yet all of them may fail to meet the evaluation threshold in terms of safety or efficiency. In the absence of explicit guidance, the negotiator is then likely to rely on similar reasoning patterns in subsequent rounds, repeatedly proposing comparable or ineffective strategies. This behavior can lead to local stagnation and reduce convergence efficiency, even when multiple candidate plans are considered.
To address this issue, we introduce ESRM Module, which performs short-term corrective reflection by analyzing low-score candidate plans and providing targeted improvement suggestions. These suggestions explicitly guide subsequent negotiation rounds, thereby improving plan quality and accelerating local consensus convergence.

\begin{figure}[t]
\centering
\includegraphics[width=0.95\columnwidth]{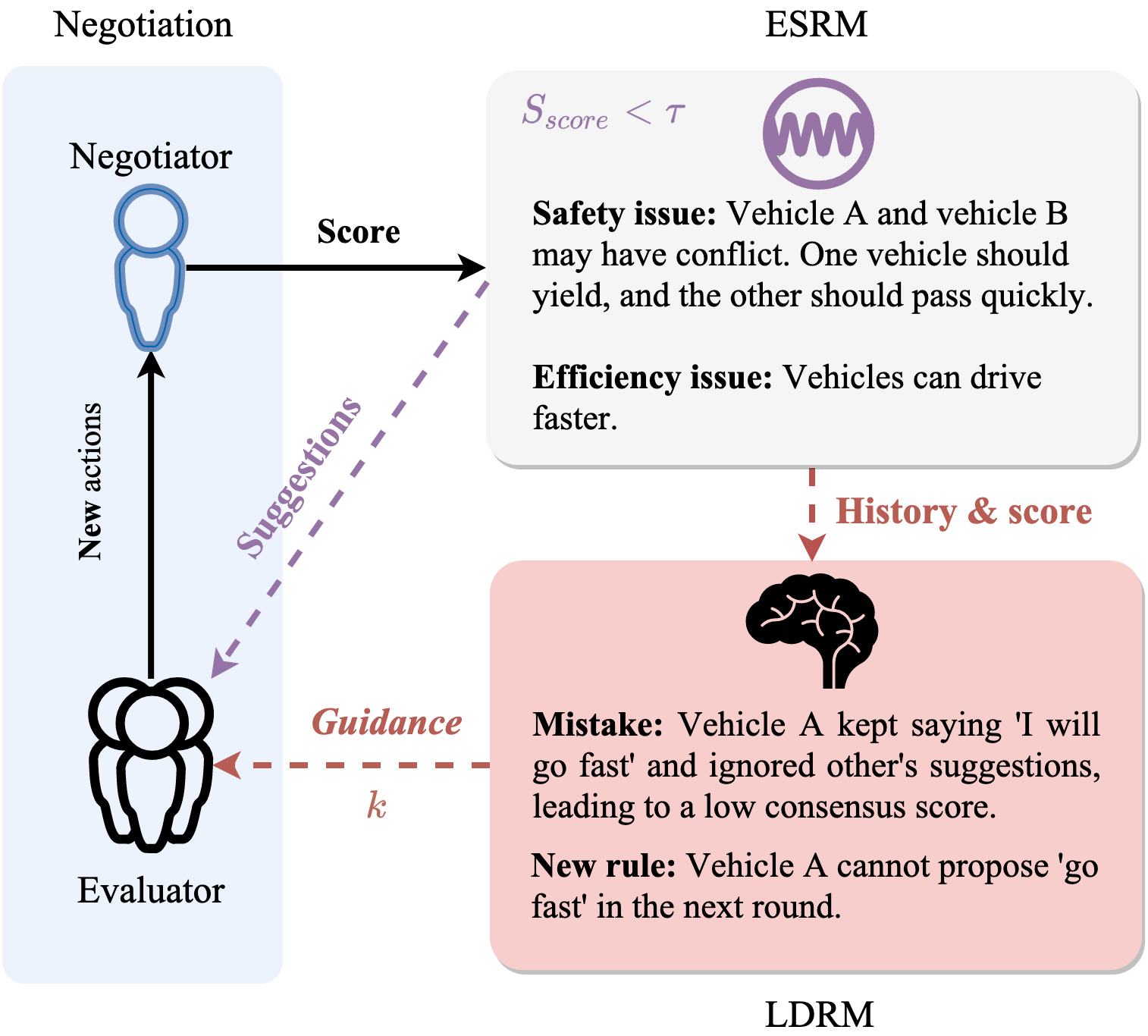} 
\caption{The detailed reasoning workflows of the ESRM Module and the LDRM Module.}
\label{Fig-reflect}
\end{figure}

As illustrated in Fig.~\ref{Fig-reflect}, during the initial negotiation round, the ESRM does not provide any feedback. The negotiator generates multiple candidate plans based on its own driving intention and the intentions of surrounding vehicles. These candidate plans are then evaluated by the evaluator.
If no candidate plan meets the predefined criteria, the ESRM analyzes the low-scoring plans and generates concise feedback from the perspectives of safety and efficiency. For example, when a plan receives a low safety score, the ESRM recommends yielding or deceleration.
When a plan receives a low efficiency score, it suggests adjusting the vehicle’s speed to improve efficiency. The feedback is incorporated into the next negotiation round as explicit guidance, steering the negotiator away from ineffective strategies and improving local consensus efficiency.

Formally, the feedback generated by the ESRM in negotiation round $r$ is expressed as:
\begin{align}
F_{\text{sugg}}^{r} = \text{LLM} \Big( f( I^{r}, S_{\text{score}}^{r} ) \Big),
\end{align}
where $I^{r}$ denotes the aggregated driving intentions in negotiation round $r$, $S_{\text{score}}^{r}$ represents the corresponding evaluation scores, including safety and efficiency.
$F_{\text{sugg}}^{r}$ is the concise feedback provided for the subsequent negotiation round.

\subsection{LLM-based Deep Reflection Module}
\label{DRM}

While the LMin and ESRM modules can effectively improve negotiation quality through multi-decision path generation and short-term feedback, they are insufficient to address a critical issue that emerges in prolonged negotiations: overconfidence and reasoning fixation. 
\color{black}
In CoLMIN, overconfidence refers to an observable negotiation behavior in which a model continues to maintain  its current decision, even after receiving feedback from the previous negotiation round indicating that the decision is inappropriate and should be adjusted.
\color{black}
This persistent overconfidence suppresses exploration of alternative solutions and gradually leads to reasoning fixation, where negotiation fails to benefit from the inherent multi-solution nature of driving decisions. As a result, the system may stagnate or converge to suboptimal outcomes despite continuous short-term refinements provided by the ESRM.

\color{black}
\color{black}

To address this issue, we introduce an LLM-based Deep Reflection Module (LDRM), which focuses on identifying and correcting long-term cognitive biases accumulated across multiple negotiation rounds.
Unlike the ESRM, which provides round-to-round refinement signals, \color{black}LDRM is activated only when negotiation persists beyond a predefined number of rounds $k$. \color{black}
The module records the negotiation history, including proposed action plans, reasoning traces, and evaluation scores, and leverages the LLM to perform holistic analysis of these records to detect persistent overconfidence and fixation patterns.
Based on this analysis, LDRM generates high-level corrective guidance to encourage exploration of alternative strategies and restore diversity in the negotiation process.

To enable systematic detection of such long-term cognitive biases, we design a set of criteria for the LDRM, including:
(i) \emph{High repetition}, where a vehicle repeatedly proposes identical or highly similar strategies across multiple rounds;
(ii) \emph{Low evaluation scores}, indicating consistently poor performance under the evaluator’s safety and efficiency metrics;
(iii) \emph{Overconfident expressions}, characterized by exaggerated or unsupported assertions in generated strategies;
and (iv) \emph{Deviation from group consensus}, where proposed strategies  diverge from the emerging collective agreement.
These criteria are encoded into the LLM prompts, allowing the model to automatically identify long-term cognitive biases when analyzing negotiation histories and to generate corresponding corrective suggestions.
As illustrated in Fig.~\ref{Fig-reflect}, LDRM is triggered as the negotiation process becomes prolonged.
The resulting guidance is fed back to the LMin module to regulate subsequent negotiation behavior, preventing vehicles from repeatedly adhering to rigid or unreasonable strategies and facilitating stable consensus formation while avoiding convergence to suboptimal solutions.

\subsection{Low-Level Control Module}

Once a consensus strategy is reached among multiple vehicles, a low-level control module is responsible for translating the negotiated high-level decision intents into executable vehicle control commands.
In this work, we adopt the low-level control design proposed in \cite{colmdriver} without modification, focusing instead on improving the upstream negotiation and decision-making process.
Specifically, bird’s-eye-view (BEV) features from multiple vehicles, along with their individual driving intents and the negotiated consensus intent, are fed into a waypoint generator to produce a sequence of geometrically feasible waypoints.
These waypoints are subsequently passed to the control module, which computes continuous control signals, including throttle, braking, and steering angle, to execute the planned trajectory.
In the simulation environment, the low-level control module operates at a high frequency and outputs control commands at each control step.
Notably, the proposed framework intervenes in the control pipeline only when intent conflicts among vehicles are detected and negotiation is required.
Under normal driving conditions without conflicts, vehicles rely solely on the low-level controller operating independently at high frequency, thereby ensuring real-time performance and overall system stability.

\section{Experiments}
In this section, we evaluate the CoLMIN framework through closed-loop simulations in the CARLA environment. We first outline the experimental setup and evaluation metrics, followed by a systematic assessment of the overall performance of our method and its key components.

\begin{table*}[]
 \caption{\color{black}Driving performance in V2Xverse simulator. Red denotes the best and black the second-best or the same results. CoLMIN performs excellently across all scenarios, especially  in driving score.\color{black}
}
\renewcommand{\arraystretch}{1.25}
\resizebox{\textwidth}{!}{
\begin{tabular}{l|cccc|cccc|cccc|cccc}
\hline
\multicolumn{17}{c}{Without Traffic Participants}     \\ \hline
                                  & \multicolumn{4}{c|}{\textbf{Total}}                                                                                                                         & \multicolumn{4}{c|}{\textbf{IC}}                                                                                                                            & \multicolumn{4}{c|}{\textbf{LM}}                                                                                                                             & \multicolumn{4}{c}{\textbf{LC}}                                                                                                                              \\
\multirow{-2}{*}{\textbf{Method}} & \textbf{DS↑}                          & \textbf{RC↑}                          & \textbf{IS↑}                         & \textbf{SR↑}                         & \textbf{DS↑}                          & \textbf{RC↑}                          & \textbf{IS↑}                         & \textbf{SR↑}                         & \textbf{DS↑}                          & \textbf{RC↑}                           & \textbf{IS↑}                         & \textbf{SR↑}                         & \textbf{DS↑}                          & \textbf{RC↑}                           & \textbf{IS↑}                         & \textbf{SR↑}                         \\ \hline
VAD (CVPR2021) \cite{single2}                              & 25.37                                 & 75.00                                 & 0.33                                 & 0.02                                 & 15.49                                 & 54.72                                 & 0.29                                 & 0.00                                 & 37.24                                 & 92.85                                  & 0.40                                 & 0.05                                 & 17.93                                 & 76.00                                  & 0.26                                 & 0.00                                 \\
UniAD (CVPR2023)  \cite{single4}                           & 35.17                                 & 88.30                                 & 0.38                                 & 0.11                                 & 37.24                                 & 91.63                                 & 0.41                                 & 0.11                                 & 42.50                                 & 84.41                                  & 0.47                                 & 0.15                                 & 12.19                                 & {\color[HTML]{333333} 90.57}           & 0.12                                 & 0.00                                 \\
TCP (NeurIPS2022) \cite{TCP}                              & 73.68                                 & 90.54                                 & 0.82                                 & 0.50                                 & 77.64                                 & 82.83                                 & {\color[HTML]{3531FF} \textbf{0.94}} & 0.50                                 & 82.18                                 & 95.18                                  & 0.86                                 & 0.70                                 & 43.52                                 & 96.30                                  & 0.45                                 & 0.00                                 \\
LMDriver (CVPR2024) \cite{lmdrive}                         & 48.83                                 & 58.02                                 & 0.85                                 & 0.20                                 & 44.72                                 & 57.94                                 & 0.79                                 & 0.17                                 & 60.88                                 & 69.43                                  & 0.86                                 & 0.30                                 & 27.96                                 & 29.70                                  & {\color[HTML]{FF0000} \textbf{0.95}} & 0.00                                 \\ \hline
CoDriving (Tpami2025) \cite{CoDriving}                        & 74.13                                 & {\color[HTML]{FF0000} \textbf{96.31}} & 0.76                                 & 0.57                                 & 66.32                                 & 90.57                                 & 0.71                                 & 0.61                                 & 96.18                                 & {\color[HTML]{3531FF} \textbf{100.00}}                                 & 0.96                                 & 0.75                                 & 36.57                                 & {\color[HTML]{FF0000} \textbf{100.00}} & 0.37                                 & 0.00                                 \\
Colmdriver (ICCV2025) \cite{colmdriver}                       & {\color[HTML]{3531FF} \textbf{88.53}} & 94.05                                 & {\color[HTML]{3531FF} \textbf{0.90}} & {\color[HTML]{3531FF} \textbf{0.80}} & {\color[HTML]{3531FF} \textbf{82.07}} & 88.78                                 & 0.86                                 & 0.72                                 & {\color[HTML]{3166FF} \textbf{98.27}} & 99.93                                  & {\color[HTML]{3531FF} \textbf{0.98}} & {\color[HTML]{FF0000} \textbf{0.92}} & {\color[HTML]{3531FF} \textbf{59.21}} & {\color[HTML]{3531FF} \textbf{82.50}}  & 0.60                                 & {\color[HTML]{3531FF} \textbf{0.50}} \\
CoLMIN(Ours)                      & {\color[HTML]{FF0000} \textbf{90.12}} & {\color[HTML]{3531FF} \textbf{94.36}} & {\color[HTML]{FF0000} \textbf{0.92}} & {\color[HTML]{FF0000} \textbf{0.81}} & {\color[HTML]{FF0000} \textbf{90.71}} & {\color[HTML]{3531FF} \textbf{93.97}} & {\color[HTML]{FF0000} \textbf{0.94}} & {\color[HTML]{FF0000} \textbf{0.78}} & {\color[HTML]{FF0000} \textbf{98.75}} & {\color[HTML]{FF0000} \textbf{100.00}} & {\color[HTML]{FF0000} \textbf{0.99}} & {\color[HTML]{3531FF} \textbf{0.90}} & {\color[HTML]{FF0000} \textbf{67.20}} & 81.12                                  & {\color[HTML]{3531FF} \textbf{0.68}} & {\color[HTML]{FF0000} \textbf{0.63}} \\ \hline
\end{tabular}
}
\label{Tab-all1}
\end{table*}

\begin{table*}[t]
\caption{\color{black}Driving performance in V2Xverse simulator with traffic participants. CoLMIN consistently achieves the best performance, demonstrating strong robustness.\color{black}}
\renewcommand{\arraystretch}{1.25}
\resizebox{\textwidth}{!}{
\begin{tabular}{lllllllllllllllll}
\hline
\multicolumn{17}{c}{With Traffic Participants}                                                                                                                                                                                       \\ \hline
\multicolumn{1}{l|}{\multirow{2}{*}{\textbf{Method}}} & \multicolumn{4}{c|}{\textbf{Total}}                       & \multicolumn{4}{c|}{\textbf{IC}}                          & \multicolumn{4}{c|}{\textbf{LM}}                           & \multicolumn{4}{c}{\textbf{LC}}      \\
\multicolumn{1}{l|}{}                        & \textbf{DS↑}   & \textbf{RC↑}   & \textbf{IS↑}  & \multicolumn{1}{l|}{\textbf{SR↑}}  & \textbf{DS↑}   & \textbf{RC↑}   & \textbf{IS↑}  & \multicolumn{1}{l|}{\textbf{SR↑}}  & \textbf{DS↑}   & \textbf{RC↑}    & \textbf{IS↑}  & \multicolumn{1}{l|}{\textbf{SR↑}}  & \textbf{DS↑}   & \textbf{RC↑}   & \textbf{IS↑}  & \textbf{SR↑}  \\ \hline
\multicolumn{1}{l|}{VAD (CVPR2021) \cite{single1}}                     & 25.18 & 75.66 & 0.31 & \multicolumn{1}{l|}{0.02} & 22.13 & 61.31 & 0.32 & \multicolumn{1}{l|}{0.00} & 35.10 & 88.23  & 0.39 & \multicolumn{1}{l|}{0.05} & 7.24  & 76.56 & 0.09 & 0.00 \\
\multicolumn{1}{l|}{UniAD (CVPR2023) \cite{single4}}                   & 37.13 & 88.71 & 0.41 & \multicolumn{1}{l|}{0.11} & 37.45 & 83.52 & 0.44 & \multicolumn{1}{l|}{0.06} & 48.29 & 91.33  & 0.52 & \multicolumn{1}{l|}{0.20} & 8.48  & 93.82 & 0.09 & 0.00 \\
\multicolumn{1}{l|}{TCP (NeurIPS2022) \cite{TCP}}                     & 74.18 & 91.21 & 0.82 & \multicolumn{1}{l|}{0.48} & 76.26 & 84.62 & 0.91 & \multicolumn{1}{l|}{0.44} & 86.59 & 95.00  & 0.91 & \multicolumn{1}{l|}{0.65} & 38.50 & 96.56 & 0.40 & 0.13 \\
\multicolumn{1}{l|}{LMDriver (CVPR2024) \cite{lmdrive}}                & 49.95 & 61.61 & 0.84 & \multicolumn{1}{l|}{0.13} & 47.65 & 59.34 & 0.81 & \multicolumn{1}{l|}{0.00} & 54.12 & 67.10  & 0.85 & \multicolumn{1}{l|}{0.20} & 44.69 & 53.02 &  {\color[HTML]{FF0000} \textbf{0.87}} & 0.25 \\ \hline
\multicolumn{1}{l|}{CoDriving (Tpami2025) \cite{CoDriving}}               & 64.50 & {\color[HTML]{FF0000} \textbf{93.58}} & 0.67 & \multicolumn{1}{l|}{0.47} & 54.64 &  {\color[HTML]{3531FF} \textbf{88.08}} & 0.59 & \multicolumn{1}{l|}{0.33} & 88.07 & 96.55  & 0.91 & \multicolumn{1}{l|}{0.73} & 27.78 &  {\color[HTML]{FF0000} \textbf{98.51}} & 0.28 & 0.13 \\
\multicolumn{1}{l|}{Colmdriver (ICCV2025) \cite{colmdriver}}              & {\color[HTML]{3531FF} \textbf{77.09}} & 92.02 &  {\color[HTML]{3531FF} \textbf{0.80}} & \multicolumn{1}{l|}{ {\color[HTML]{3531FF} \textbf{0.63}}} &  {\color[HTML]{3531FF} \textbf{63.06}} & 82.55 & 0.70 & \multicolumn{1}{l|}{ {\color[HTML]{3531FF} \textbf{0.44}}} &  {\color[HTML]{3531FF} \textbf{94.00}} &  {\color[HTML]{3531FF} \textbf{100.00}} &  {\color[HTML]{3531FF} \textbf{0.94}} & \multicolumn{1}{l|}{ {\color[HTML]{3531FF} \textbf{0.85}}} &  {\color[HTML]{3531FF} \textbf{66.38}} &  {\color[HTML]{3531FF} \textbf{93.41}} & 0.68 &  {\color[HTML]{3531FF} \textbf{0.50}} \\
\multicolumn{1}{l|}{CoLMIN (Ours)}                  & {\color[HTML]{FF0000} \textbf{78.93}}     & {\color[HTML]{3531FF} \textbf{92.79}}   &  {\color[HTML]{FF0000} \textbf{0.80}}    & \multicolumn{1}{l|}{{\color[HTML]{FF0000} \textbf{0.67}}}    & {\color[HTML]{FF0000} \textbf{66.47}}     & {\color[HTML]{FF0000} \textbf{89.17}}     &  {\color[HTML]{FF0000} \textbf{0.70}}    & \multicolumn{1}{l|}{{\color[HTML]{FF0000} \textbf{0.53}}}    & {\color[HTML]{FF0000} \textbf{94.11}}      & {\color[HTML]{FF0000} \textbf{100.00}}       & {\color[HTML]{FF0000} \textbf{0.94}}     & \multicolumn{1}{l|}{{\color[HTML]{FF0000} \textbf{0.85}} }    & {\color[HTML]{FF0000} \textbf{67.45}} 
     & 82.45     &  {\color[HTML]{3531FF} \textbf{0.68}}    & {\color[HTML]{FF0000} \textbf{0.63}}   \\ \hline
\end{tabular}
}
\label{Tab-all2}
\end{table*}

\subsection{Experimental Setup}

We conduct closed-loop evaluations on the CARLA simulation platform \cite{carla}, with scenario data generated using the V2Xverse simulator.
A total of 10 representative multi-vehicle scenarios are designed to emulate realistic cooperative autonomous driving environments.
In each scenario, multiple  vehicles are evaluated, while additional traffic participants, including pedestrians and cyclists, are introduced as dynamic or static obstacles.
All simulations are executed with a control frequency of 5~Hz.
Regarding model configuration, CoLMIN employs a Qwen2.5-3B model fine-tuned with LoRA \cite{lora} for the LMin Module , and a Qwen2.5-7B model for the LDRM Module , balancing reasoning capability and computational efficiency.
\color{black}
The negotiation termination threshold is set to $\tau = 0.85$, which is empirically determined through preliminary experiments. \color{black}The maximum number of negotiation rounds is set to $R_{max}=3, k = 2$. \color{black} Following the safety-first principle in autonomous driving, the weights of the overall score, i.e., $w_S$, $w_E$, and $w_C$, are set to 0.5, 0.2, and 0.3, respectively. This weighting scheme assigns the highest priority to safety, while also taking driving efficiency and consensus formation into account. All these hyperparameters are kept fixed across all evaluation scenarios. 
All compared methods are evaluated under the same CARLA scenarios, perception inputs, and low-level controller. The low-level controller is directly adopted from \cite{colmdriver} with the same weights and kept unchanged across all methods. \color{black}
All experiments are conducted on a computing platform equipped with four NVIDIA A100 GPUs.

\color{black}
To comprehensively evaluate system performance, we adopt four primary metrics: \textbf{Driving Score (DS)}, \textbf{Route Completion (RC)},\textbf{Infraction Score (IS)}  and \textbf{Success Rate (SR)}. \color{black}
Driving Score (DS) is the primary ranking metric. It reflects both task progress and driving safety by combining route completion with penalties for safety violations.
Route Completion (RC) measures the percentage of the predefined planned route that is successfully completed by the vehicles under test. It indicates how much progress the vehicles make toward completing the assigned driving task.
Infraction Score (IS) evaluates driving safety. It starts at 1.0 for each task and is reduced when collisions occur, using a predefined multiplicative discount factor. This metric captures the overall safety performance of all test vehicles.
Success Rate (SR) is the percentage of tasks completed with a full Driving Score. It measures the stability and reliability of the system across different test scenarios.
Together, these metrics provide a comprehensive evaluation of navigation performance, task completion, and safety in multi-vehicle driving scenarios.
\color{black}

\begin{table}[]
 \caption{Driving performance on the V2Xverse simulation platform. Red denotes the best and black the second-best results.
}
\renewcommand{\arraystretch}{1.25}
\begin{tabular}{llll|llll}
\hline
\textbf{Single} & \textbf{LMin} & \textbf{ESRM} & \textbf{LDRM} & \textbf{DS}                           & \textbf{RC}                           & \textbf{IS}                          & \textbf{SR}                          \\ \hline
       &       &     &      & 47.64                        & {\color[HTML]{FF0000} 96.43} & 0.485                       & 0.13                        \\
\checkmark      &       &     &      & 70.22                       & 87.03                        & 0.77                        & 0.54                        \\
\checkmark      & \checkmark     &     &      & 77.29                        &  95.22                    & 0.78                     &  0.65                   \\
\checkmark      & \checkmark     & \checkmark   &      & 90.02                        & 94.72                        & 0.90                        & 0.78                        \\
\checkmark      & \checkmark     & \checkmark   & \checkmark    & {\color[HTML]{FF0000} 90.12} & 94.36                        & {\color[HTML]{FF0000} 0.92} & {\color[HTML]{FF0000} 0.80} \\ \hline
\end{tabular}
\label{Tab-ab}
\end{table}

\begin{figure}[t]
\centering
\includegraphics[width=0.9\columnwidth]{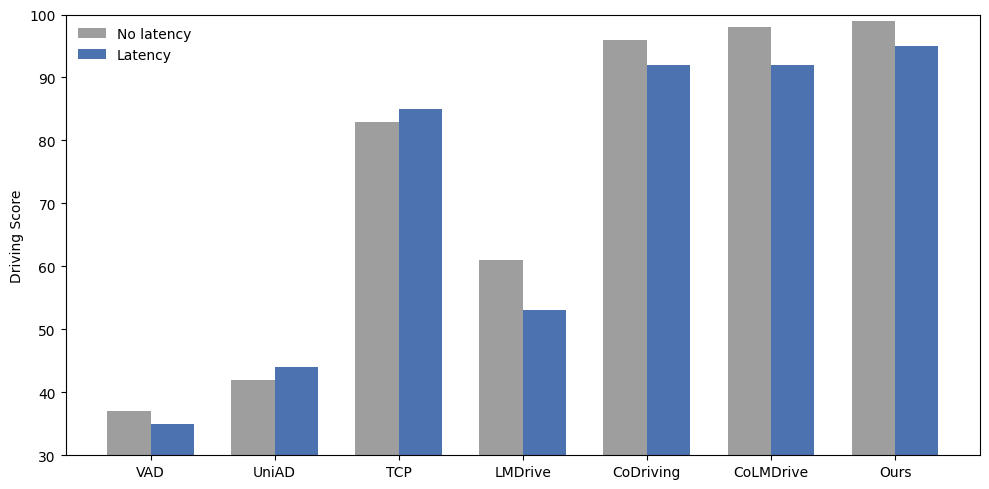} 
\caption{Comparison of driving performance with vs. without latency.}
\label{latency}
\end{figure}

\subsection{Quantitative Results of Closed-loop driving}
To validate the overall performance of CoLMIN, we compare it against a series of advanced closed-loop autonomous driving baselines, including VAD \cite{single2}, TCP \cite{single4}, UniAD \cite{UniAD}, CoDriving \cite{CoDriving}, and CoLMDriver \cite{colmdriver}. 
\color{black}
We also include comparisons with another VLM-based method, LMDriver \cite{lmdrive}. 
Specifically, CoDriving uses LLaMA2-7B, whereas CoLMDriver adopts Qwen2.5-3B as the LLM negotiator. \color{black}
For a fair comparison, we do not include optimization-based cooperative planning approaches in the comparison. 
Table~\ref{Tab-all1} summarizes the overall performance across all test scenarios, along with the results in Intersection Crossing (IC), Lane Merging (LM), and Lane Changing (LC). CoLMIN achieves the highest driving score in all interactive scenarios, outperforming single-agent baselines and demonstrating the clear advantage of collaborative negotiation in resolving conflicts.
Compared with the current best-performing method, CoLMDriver, CoLMIN achieves a 1.59\% improvement in overall driving score (90.12\% vs. 88.53\%). The gain is even more significant in the challenging LC scenario, \color{black}where CoLMIN surpasses CoLMDriver by 7.99\% (67.20\% vs. 59.21\%). \color{black}These results indicate that exploiting the multi-modality of intentions helps agents generate more effective conflict-resolution strategies, thereby improving the safety and robustness of cooperative autonomous driving.
Other baselines show notable limitations.  Despite having the best safety score, LMDrive has a very low Route Completion (RC) and Success Rate (SR)  because its overly conservative yielding rules cause all vehicles to stall. 
TCP achieves the highest safety score (IS) in the IC scenario; however, its low Success Rate (SR) reflects frequent collisions across different scenes. CoDriving obtains the highest Route Completion (RC) but exhibits elevated collision risks.
In contrast, CoLMIN leverages multi-decision path based negotiation and reflection mechanisms to simultaneously enhance safety and efficiency. It effectively prevents collisions caused by intention conflicts while avoiding deadlocks induced by excessive yielding, thereby enabling smoother and more reliable cooperative autonomous driving.

To further approximate real-world conditions, we additionally evaluate our system in a closed-loop environment populated with dynamic traffic participants, including pedestrians and cyclists. The results under these more challenging settings are reported in Table \ref{Tab-all2}.
Compared with the results in Table \ref{Tab-all1}, the introduction of traffic participants inevitably leads to performance degradation due to increased interaction complexity and uncertainty. Nevertheless, CoLMIN consistently maintains strong performance and demonstrates superior robustness in these dynamic and highly interactive environments.

\begin{figure*}[t]
\centering
\includegraphics[width=2\columnwidth]{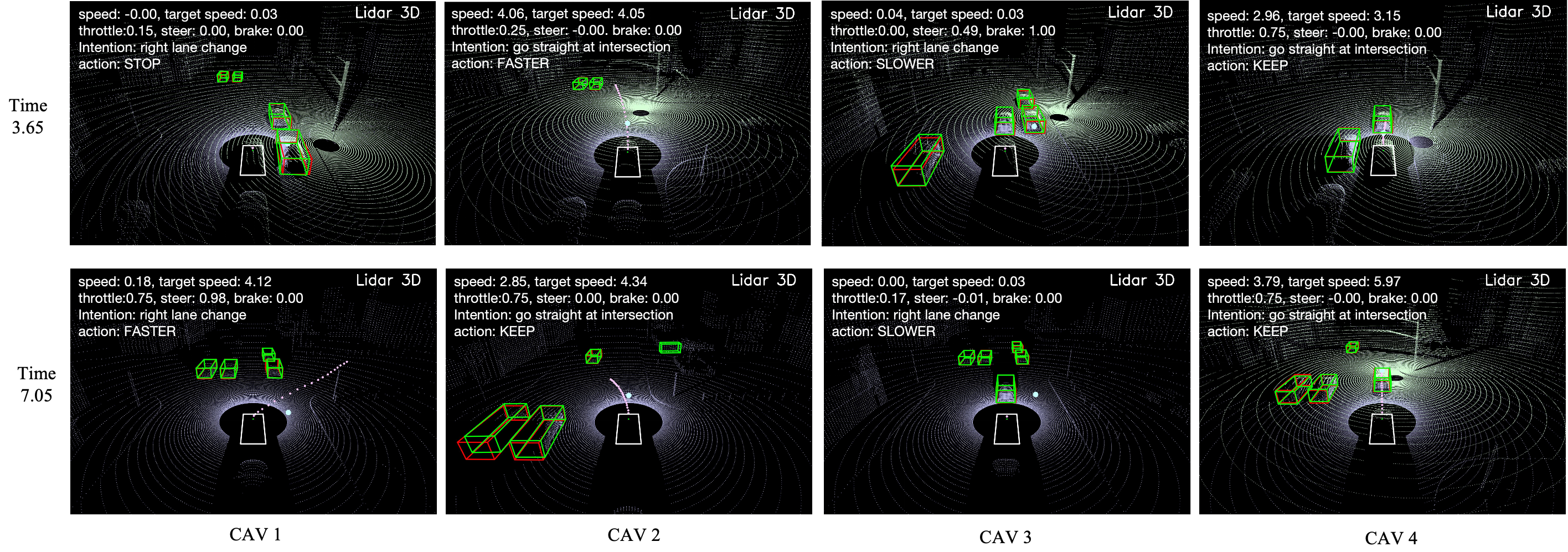} 
\caption{Visualization of CoLMIN in the Cross Change scenario. Four vehicles negotiate based on their individual states and execute a feasible joint action plan. Context slightly paraphrased for clarity. CoLMIN derives actions that are consistent with the current traffic context while ensuring collision-free coordination.}
\label{Fig-output}
\end{figure*}

\subsection{Qualitative evaluation}
Fig. \ref{Fig-output} illustrates a traffic scenario in which four vehicles negotiate on the same road. CAV 1 and 3 intend to change lanes to the right, while CAV 2 and CAV 4 proceed straight. The white outlines represent the vehicles themselves. Following the original planned actions, these vehicles would have collided at a certain point. Upon detecting this conflict, our method initiates multiple negotiation rounds. The first row of Fig. \ref{Fig-output} shows the traffic situation around 3 seconds into the simulation: through negotiation, CAV 1 decides to stop, CAV 3 slows down behind it, CAV 2 quickly passes the intersection, and CAV 4 follows behind. By the 7th second, CAV 2 and 4 have cleared the intersection; at this time, Vehicle 1 accelerates to merge right, while CAV 3 continues to move slowly, maintaining a safe distance from the vehicle ahead. This example demonstrates how our method enables driving behaviors in complex traffic scenarios and enhances overall traffic safety through multi-agent decision-level coordination. 

 \begin{figure}[t]
 \begin{center}
 \subfigure{
\includegraphics[width=0.46\linewidth]{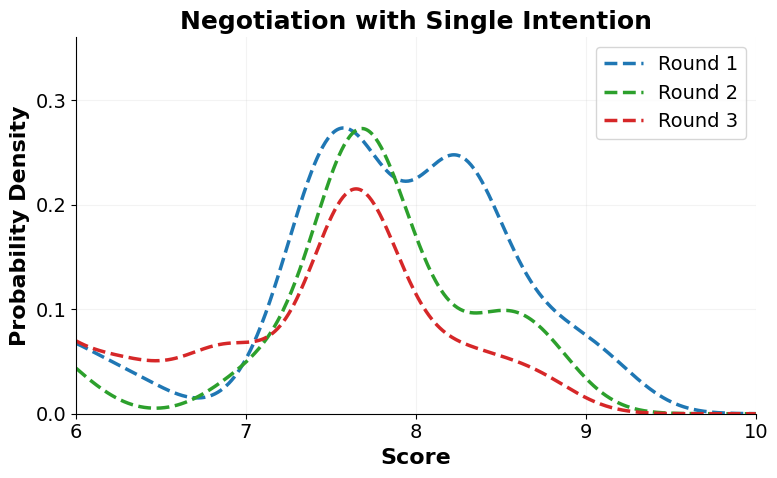} }
\hspace{-1mm}
 \subfigure{
\includegraphics[width=0.46 \linewidth]{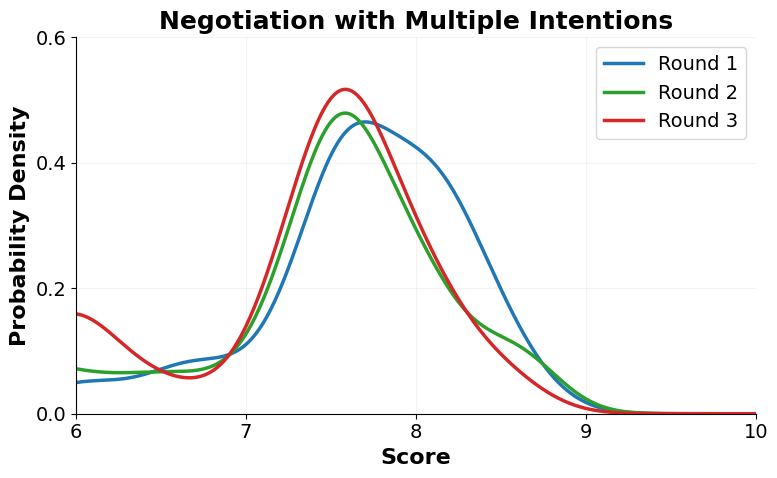} }
 \end{center}
  \vspace{-0.3cm}
 \caption{Comparison of Negotiation Convergence Under Single-Decision path based and Multi-Decision path based Paradigms.
 }
 \label{fig-con}
\end{figure}

\begin{table}[t]
\centering
\caption{\color{black}Backbone controlled ablation study.\color{black}}
\label{tab:backbone_ablation}
\begingroup
\setlength{\tabcolsep}{3pt}
\renewcommand{\arraystretch}{1.2}
\begin{tabular}{@{}lcccc@{}}
\hline
Variant & LMin & ESRM & LDRM & \shortstack{\rule{0pt}{2.5ex}Driving Score\\(LC)} \\
\hline
CoLMDriver-Qwen2.5-3B & -- & -- & -- & 59.21 \\
CoLMDriver-Qwen2.5-7B & -- & -- & -- & 59.15 \\
CoLMIN w/o LDRM & 3B & 3B & -- & 62.30 \\
CoLMIN w/ LDRM-Qwen2.5-3B & 3B & 3B & 3B & 66.00 \\
CoLMIN w/ LDRM-Qwen2.5-7B & 3B & 3B & 7B & \textbf{67.20} \\
\hline
\end{tabular}
\label{tab-ldrm}
\endgroup
\end{table}

\begin{table}[t]
\centering
\caption{\color{black}Average total inference time in the closed-loop setting.\color{black}}
\renewcommand{\arraystretch}{1.2}
\begin{tabular}{l|llll}
\hline
Inference Time (s) & \textbf{Total↓} & \textbf{IC↓}    & \textbf{LM↓}   & \textbf{LC↓}    \\ \hline
Colmdriver \cite{colmdriver}        &   31.74    &    32.93   &  26.80     &   41.44    \\
CoLMIN             & 29.67 & 31.26 & 24.97 & 37.83 \\ \hline
\end{tabular}
\label{tab-time}
\end{table}

\begin{table}[t]  
	\centering
	\caption{\color{black}Representative failure cases from closed-loop evaluation.\color{black}}
	\setlength{\tabcolsep}{2pt}  
	\renewcommand{\arraystretch}{1.15}  
	\begin{tabular}{lcccc}
		\hline
		Failure Type & Driving Score & Route Score & Penalty & Main Infraction \\
		\hline
		Collision failure & 60.0 & 100.0 & 0.6 & Vehicle collision \\
		Timeout failure & 3.988 & 3.988 & 1.0 & Route timeout \\
		Partial timeout & 66.338 & 66.338 & 1.0 & Route timeout \\
		Timeout+stop & 53.959 & 67.449 & 0.8 & Timeout+stop \\
		\hline
	\end{tabular}
	\label{tab:failure_case}
\end{table}

\begin{figure}[h]
\centering
\includegraphics[width=1\columnwidth]{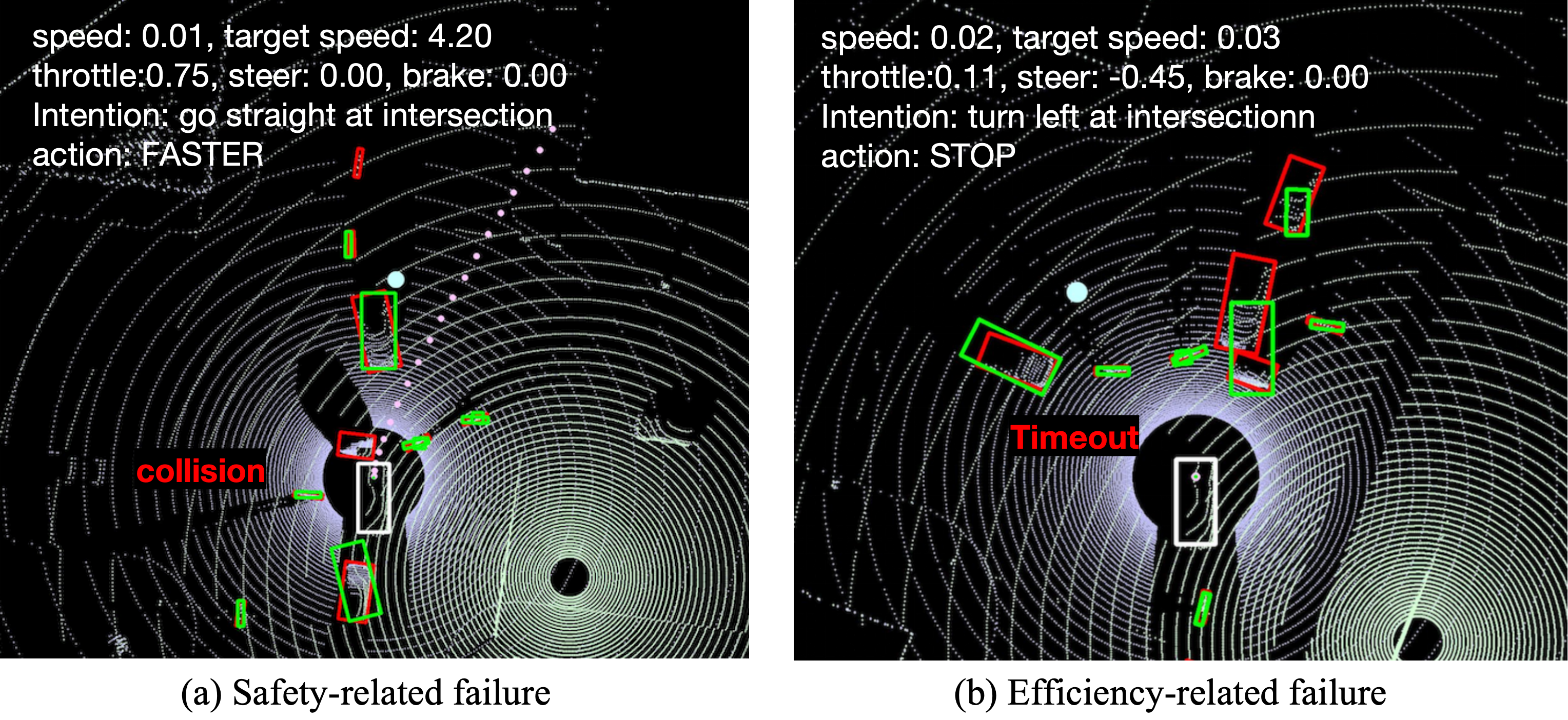} 
\caption{\color{black}Visualization experiment of failure cases.\color{black}}
\label{failure_case}
\end{figure}

\begin{table}[t]
	\centering
	\caption{\color{black}Effect of number of candidate intents on performance and efficiency.\color{black}}
	\label{tab:candidate_intents}
	\setlength{\tabcolsep}{3pt}
	\renewcommand{\arraystretch}{1.15}
	\begin{tabular}{c|ccccc}
		\hline
		Candidate Intents & DS & RC & IS & SR & Inference Time (s) \\
		\hline
		1 & 52.86 & 79.89 & 0.63 & 0.2 & 37.1 \\
		2 & 66.87 & 79.67 & 0.87 & 0.3 & 30.4 \\
		3 & 80.07 & 86.64 & 0.80 & 0.8 & 34.2 \\
		4 & 80.25 & 83.15 & 0.81 & 0.8 & 33.0 \\
		\hline
	\end{tabular}
\label{candidate}
\end{table}

\begin{table}[h]
	\centering
\caption{\color{black}Process-level analysis of negotiation convergence and efficiency, averaged over the evaluated negotiation samples. 
``Overconf.'', ``Conv.'', and ``Nego.'' denote overconfidence, convergence, and negotiation, respectively.\color{black}}
    \label{tab:process_analysis}
    \small
    \setlength{\tabcolsep}{3pt}
    \renewcommand{\arraystretch}{1.15}
    \resizebox{\linewidth}{!}{
    \begin{tabular}{l|ccccc}
        \hline
        \textbf{Method} 
        & Overconf.$\downarrow$ 
        & Rounds $\downarrow$ 
        & Conv. Rate $\uparrow$ 
        & Nego. Time 
        & Tokens $\downarrow$ \\
        \hline
        Baseline 
        & 2.08 
        & 2.52 
        & 33.90\% 
        & 8.06s 
        & 4487 \\
        Ours 
        & \textbf{1.37} 
        & \textbf{1.90} 
        & \textbf{86.25\%} 
        & 9.51s 
        & 5053 \\
        \hline
    \end{tabular}
    }
\end{table}

\subsection{Ablation Study}
To evaluate the contribution of each component, we conduct a comprehensive ablation study by selectively removing individual modules while keeping the rest of the system unchanged. We then further investigate system-level properties, including latency performance and consensus convergence behavior, to provide a more complete understanding of the proposed framework.

\textbf{System Component Ablation.} We evaluated the impact of different system modules on overall performance in Table \ref{Tab-ab}. The first row represents the results without any negotiation. Although each vehicle attempts to complete its own driving route, severe collisions occur, resulting in a very low Success Rate of only 0.13, and the overall driving performance is minimal. The second row incorporates the negotiation module but considers only a single behavior intent. Here, overall driving performance improves significantly, and the safety score rises to 0.54, demonstrating the effectiveness of multi-agent negotiation strategies. Building on this, the introduction of the multi-decision path based negotiation mechanism further enhances performance: the Driving Score increases by 7.07\% (77.29\% vs. 70.22\%), and the Route Completion improves by 8.19\% (95.22\% vs. 87.03\%). When both the ESRM and LDRM Modules are integrated into the system, feedback proves highly beneficial, boosting overall performance by over 10\%. Notably, the LDRM Module reduces collision occurrences, resulting in improvements in both the Infraction Score and Success Rate.

\textbf{Latency performance.}
Fig. \ref{latency} compares the performance of existing methods in the Lane Merging scenario under both ideal computing conditions (no inference latency) and conditions with inference latency. As shown in the figure, most baselines experience varying degrees of performance degradation when latency is introduced. For example, LMDrive, CoDriving, and CoLMDriver show decreases of approximately 7\%, 4\%, and 6\%, respectively. In contrast, our CoLMIN exhibits only a 3\% drop and still maintains a score above 90. This demonstrates the strong stability of our method, which is able to preserve high performance even in the presence of inference latency.

\textbf{Consensus Convergence.} Fig. \ref{fig-con} illustrates the distribution of negotiation quality scores for both single-decision and multi-decision path negotiation strategies over three iterative rounds.
When negotiation is updated using only a single intent, the score distribution exhibits noticeable random fluctuations across rounds, with poor convergence and generally low score concentrations. This indicates that single-intent negotiation struggles to consistently improve strategy quality through iterative interactions.
In contrast, the multi-decision path negotiation strategy explores a richer intent space in each round and steadily refines the strategy quality under the guidance of the evaluator mechanism. The results show that the score distributions under multi-decision path negotiation are significantly smoother and more stable, reflecting stronger robustness and a clearer convergence trend.
Therefore, the proposed multi-decision path negotiation mechanism effectively reduces negotiation oscillation and accelerates consensus formation.

\color{black}

\textbf{Inference time.} To further assess the computational efficiency of our framework, we evaluated the total time required to complete the closed-loop negotiation and reasoning process in three different traffic scenarios, with the unit of time measured in seconds. As shown in Table \ref{tab-time}, CoLMIN achieves an average total inference time of 29.67 s, which is lower than that of CoLMDriver, demonstrating improved efficiency despite incorporating  multi-decision path  reasoning and dual reflection modules. In particular, CoLMIN is faster across all scenarios, with the largest improvement observed in the LC scenario, where the average time is reduced from 41.44 s to 37.83 s. These results indicate that our proposed method not only enhances driving performance but also maintains competitive computational efficiency in closed-loop settings. 
\color{black}
Noted that,  the reported inference time refers to the complete high-level LLM-based negotiation and decision-making process, rather than the inference latency of the low-level real-time controller.
In practical deployment, CoLMIN performs deep reasoning at a low frequency, while the low-level controller operates at a high frequency to ensure real-time control.
\color{black}

\textbf{The effect of model capacity.}
To decouple the contribution of the proposed framework from the effect of backbone model capacity, we conducted a backbone-controlled ablation study, as shown in Table~\ref{tab:backbone_ablation}. First, replacing the backbone of CoLMDriver from Qwen2.5-3B to Qwen2.5-7B does not improve performance, indicating that simply using a larger backbone model does not necessarily lead to better driving performance.
Second, CoLMIN with LDRM-Qwen2.5-3B achieves a driving score of 66.00\%, which is 3.70\% higher than CoLMIN without LDRM. This demonstrates that the performance gain mainly comes from the proposed deep reflection mechanism rather than from the use of a larger model.
Finally, the full CoLMIN model with Qwen2.5-7B for LDRM achieves the best score of 67.20\%, an improvement of 1.20\% over the 3B-LDRM variant. This suggests that the 7B model further enhances the reliability of the deep reflection process.

\textbf{The effect of the number of candidate intents.} We have added an ablation study conducted on five complex scenarios to analyze the effect of the number of candidate intents on performance and efficiency. The results are shown in Table~\ref{candidate}. The results indicate that increasing the number of candidate intents improves the driving score, route completion, and success rate, while inference time remains largely stable. When the number of candidate intents reaches four, performance essentially plateaus. This confirms that the performance gains are primarily attributable to the multi-decision-path mechanism rather than longer generation time or increased token usage.

\textbf{Failure Case from closed-loop evaluation.}
We further analyze representative failure cases from closed-loop evaluation, as summarized in Table~\ref{tab:failure_case}. We identify two main failure modes. 
First, safety-related failures occur when the vehicle completes the route but collides at conflict points. For example, in the collision failure case, as illustrated in Fig. \ref{failure_case}(a), the Route Score is 100.0, but the Driving Score drops to 60.0 due to a vehicle collision and the corresponding penalty of 0.6. 
Second, as illustrated in Fig. \ref{failure_case}(b), efficiency-related failures occur when the vehicle fails to complete the route on time or only partially completes it. For example, the timeout failure obtains a Driving Score and Route Score of 3.988, while the partial timeout case obtains 66.338. In the ``timeout + stop sign infraction'' case, the Driving Score is 53.959 and the Route Score is 67.449.

\textbf{Controlled Process-Level Analysis.}
We conduct a controlled process-level comparison to examine whether the proposed method improves the negotiation process itself. 
As shown in Table~\ref{tab:process_analysis}, our method reduces the average number of negotiation rounds from 2.52 to 1.90 and improves the convergence rate from 33.90\% to 86.25\%. This demonstrates that the proposed method significantly increases the probability of reaching a valid consensus rather than merely terminating the negotiation earlier. Meanwhile, the average number of overconfident vehicles decreases from 2.08 to 1.37, indicating that our method mitigates overconfident decisions.
In terms of efficiency, the average token usage increases from 4487 to 5053, and the average negotiation time increases from 8.06 s to 9.51 s due to the additional reflection step. These results show that our method introduces moderate computational overhead while achieving more stable negotiation convergence.

\section{Discussion on Real-world Applicability.}
CoLMIN is evaluated in CARLA-based closed-loop simulation, following the common practice of recent cooperative autonomous driving studies. 
However, simulation-based evaluation cannot fully reflect real-world driving conditions. Practical deployment may involve perception noise, localization errors, communication latency, diverse human driving behaviors, and differences in road structures or traffic rules. In addition, real-world validation of multi-agent cooperative decision-making is costly and difficult. Many real-world driving datasets are collected in an open-loop manner, therefore, these datasets cannot directly support closed-loop testing of interactive multi-vehicle negotiation.
Thus, the current experiments mainly validate the feasibility of the proposed negotiation mechanism in closed-loop simulation. In future work, we will extend the evaluation to real-world datasets, real-world scenario replay, and world-model-based simulation platforms to further assess the generalization and practical applicability of CoLMIN.

\color{black}
\textbf{Quantitative validation of the evaluator.}
To evaluate the accuracy and reliability of LMin, we conducted a quantitative validation. Specifically, we compared the scores generated by the LMin evaluator with the actual scores obtained from closed-loop deployment. As shown in Table~\ref{tab:lmin_evaluator_validation}, the evaluator scores are very close to the objective deployment scores, with absolute gaps of 0.031, 0.043, and 0.059 for safety, efficiency, and consensus, respectively. These small gaps indicate that the scoring results provided by the LLM-based evaluator are generally consistent with the actual closed-loop performance.

\begin{table}[t]
	\centering
\caption{\color{black}Validation of the LLM-based evaluator on the final negotiation plans. LMin Score denotes the average score produced by the evaluator; Objective Ref. Score denotes the actual rollout score; Abs. Gap denotes their absolute difference.\color{black}}
	\label{tab:lmin_evaluator_validation}
	\begin{tabular}{lccc}
		\hline
		\textbf{Criterion} 
		& \textbf{LMin Score} 
		& \textbf{Objective Ref. Score} 
		& \textbf{Abs. Gap $\downarrow$} \\ \hline
		Safety 
		& 0.928 & 0.959 & 0.031 \\
		
		Efficiency 
		& 0.198 & 0.155 & 0.043 \\
		
		Consensus 
		& 0.851 & 0.910 & 0.059 \\ \hline
	\end{tabular}
\end{table}
\color{black}
\section{CONCLUSION} \label{6}

This paper presents CoLMIN, a multi-vehicle cooperative autonomous driving framework that achieves stable decision consensus through multi-decision path negotiation and reflective reasoning, aiming to improve safety in complex traffic environments.
CoLMIN comprises three main components: the LMin module, which generates and evaluates multiple candidate driving intentions; the ESRM module, which provides feedback on underperforming decisions to refine negotiation outcomes and accelerate consensus formation; and the LDRM module, which analyzes negotiation history to identify cognitive fixation, enhancing plan diversity and preventing the system from being trapped in suboptimal solutions. 
\color{black}Extensive  experiments conducted in the CARLA simulation environment demonstrate that CoLMIN delivers outstanding performance in complex interactive driving scenarios.\color{black} In particular, it exhibits significant advantages in highly challenging situations, highlighting the potential of our approach for cooperative autonomous driving. In future work, we plan to explore  more complex, interactive scenarios within a world model to further enhance the safety and reliability of autonomous driving.

\section*{Acknowledgments}

This work was supported by the National Natural Science Foundation of China under Grant No. 62441617, the Postdoctoral Fellowship Program and China Postdoctoral Science Foundation under Grant No. 2024M764093, Grant No. BX20250485 and No.2026M791776, the Beijing Natural Science Foundation under Grant No. 4254100, and by Beijing Advanced Innovation Center for Future Blockchain and Privacy Computing, and by Fundamental Research Funds for the Central Universities (No.300102376106).

\bibliographystyle{IEEEtran}
\bibliography{sample-base}

\begin{IEEEbiography}[{\includegraphics[width=1in,height=1.25in,clip,keepaspectratio]{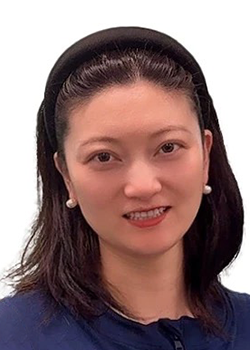}}]{Zhe Huang}
received her Ph.D. degree from the School of Information, Renmin University of China, in 2025. She obtained her M.S. degree from the Institute of Computing Technology, Chinese Academy of Sciences, and previously conducted research at the University of Wollongong, NSW, Australia. She is currently an Assistant Professor with the School of Future Transportation, Chang’an University. Her research interests include multi-agent collaboration, computer vision,  autonomous driving and multi-modal large language models.
\end{IEEEbiography}

\begin{IEEEbiography}[{\includegraphics[width=1in,height=1.25in,clip,keepaspectratio]{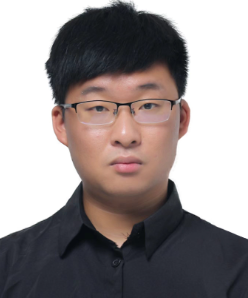}}]{Zhaoxin Fan}
received his Ph.D. degree from the School of Information, Renmin University, China in 2024. He has also conducted research at Carnegie Mellon University and  Hong Kong University of Science and Technology. He is currently an Assistant Professor in the Institute of Artificial Intelligence, Beihang University. His research interests include multi-modal large language models (LLMs), computer vision, and embodied AI.
\end{IEEEbiography}

 \begin{IEEEbiography}[{\includegraphics[width=1in,height=1.25in,clip,keepaspectratio]{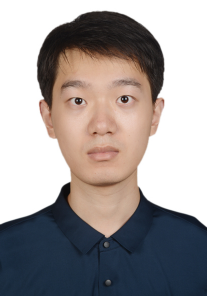}}]{Shuo Wang}
received his bachelor’s degree from the School of Information, Renmin University of China, where he is currently pursuing a Ph.D. degree. He has published multiple first-author papers in international journals and conferences, including but not limited to CVPR, NeurIPS, ICCV, ACM MM, AAAI, and TIP. His research interests include vision-language navigation, simultaneous localization and mapping, collaborative perception, and 3D reconstruction.

\end{IEEEbiography}

 \begin{IEEEbiography}[{\includegraphics[width=1in,height=1.25in,clip,keepaspectratio]{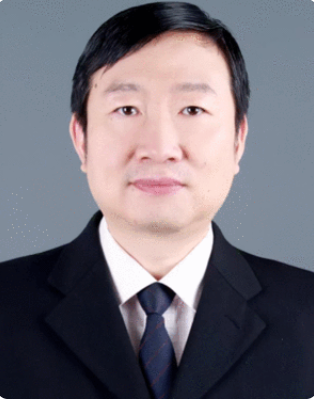}}]{Wenjun Wu} (Member, IEEE) received the Ph.D. degree in computer science from Beihang University, Beijing, China, in 2001. From 2002 to 2010, he was a Research Scientist with Indiana University, Bloomington, IN, USA, and the Argonne National Laboratory, The University of Chicago, Chicago, IL, USA. He is currently a Full Professor with the School of Artificial Intelligence, Beihang University. He has published more than 180 academic papers in international journals and conferences, and led in more than 30 national key projects, including the National Key Research and Development Program of China, NSFC Key Project. His research interests include multi-agent reinforcement learning, swarm intelligence, crowdsourcing, cloud computing, and AI for science. 
\end{IEEEbiography}

 \begin{IEEEbiography}[{\includegraphics[width=1in,height=1.25in,clip,keepaspectratio]{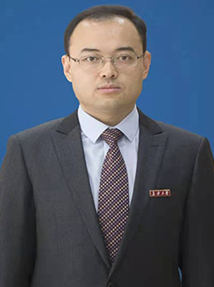}}]{Xuan Zhao}
received the B.S. degree from Chang'an University, Xi'an, China, in 2003, and the Ph.D. degree from Chang'an University, Xi'an, China, in 2009. He was a Visiting Scholar with the University of Michigan, USA, in 2017. He is currently a Professor and the Dean of the School of Automobile, Chang'an University. His research interests include control, testing, and evaluation technologies for intelligent driving vehicles.
\end{IEEEbiography}

 \begin{IEEEbiography}[{\includegraphics[width=1in,height=1.25in,clip,keepaspectratio]{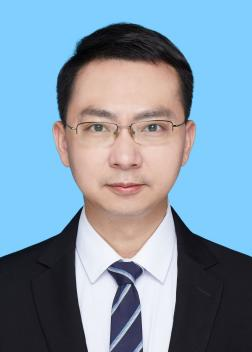}}]{Min Liu}  (Member, IEEE) received the bachelor’s degree from Peking University, Beijing, China, in 2004, and the Ph.D. degree in electrical engineering from the University of California, Riverside, Riverside, CA, USA, in 2012. He was a Research Scientist with the University of California, Santa Barbara, Santa Barbara, CA, USA. He is currently a Professor with the College of Electrical and Information Engineering, Hunan University, Changsha, China. His research interests include computer vision and image processing. Dr. Liu is an Associate Editor of IEEE Transactions on Neural Networks and Learning Systems.

\end{IEEEbiography}

\vfill
\end{document}